\documentclass[runningheads]{llncs}
\usepackage{graphicx}
\usepackage{amsmath}
\usepackage{xcolor}
\usepackage{subcaption}
\usepackage{tabularx}
\usepackage{booktabs}

\usepackage{siunitx}
\usepackage{microtype}

\usepackage{acronym}

\usepackage[hyphens]{url}
\usepackage[pagebackref=true,breaklinks=true]{hyperref}
\hypersetup{
     colorlinks,
     linkcolor={red!75!black},
     citecolor={blue!75!black},
     urlcolor={blue!75!black},
}
\usepackage[capitalise]{cleveref}

\usepackage[inline]{enumitem}
\setlist*[enumerate]{label=(\arabic*)}

\usepackage{xspace}
\makeatletter
\DeclareRobustCommand\onedot{\futurelet\@let@token\@onedot}
\newcommand{\@onedot}{\ifx\@let@token.\else.\null\fi\xspace}
\makeatother

\newcommand{\ie}{i.\,e.,\xspace}
\newcommand{\eg}{e.\,g.,\xspace}

\newcommand{\cf}{cf\onedot}

\newacro{cnn}[CNN]{Convolutional Neural Network}
\newacro{dnn}[DNN]{Deep Neural Network}

\begin{document}
\title{A Fair Evaluation of Various Deep Learning-based Document Image Binarization Approaches}
\author{Richin Sukesh\orcidID{0000-0002-1845-820X} \and 
Mathias Seuret\orcidID{0000-0001-9153-1031} \and
Anguelos Nicolaou\orcidID{0000-0003-3818-8718}  \and
Martin Mayr\orcidID{0000-0002-3706-285X} \and
Vincent Christlein\orcidID{0000-0003-0455-3799}
}
\authorrunning{R.\ Sukesh et al.}
\titlerunning{A Fair Evaluation of Deep Learning-based Binarization Methods}
\institute{Friedrich-Alexander-Universität Erlangen-Nürnberg, Erlangen, Germany\\
\email{richin.sukesh@fau.de}}
\maketitle              %
\begin{sloppypar}
\begin{abstract}
Binarization of document images is an important pre-processing step in the field of document analysis. Traditional image binarization techniques usually rely on histograms or local statistics to identify a valid threshold to differentiate between different aspects of the image. Deep learning techniques are able to generate binarized versions of the images by learning context-dependent features that are less error-prone to degradation typically occurring in document images.
In recent years, many deep learning-based methods have been developed for document binarization. But which one to choose? 
There have been no studies that compare these methods rigorously.
Therefore, this work focuses on the evaluation of different deep learning-based methods under the same evaluation protocol. 
We evaluate them on different Document Image Binarization Contest (DIBCO) datasets and obtain very heterogeneous results. 
We show that the DE-GAN model was able to perform better compared to other models when evaluated on the DIBCO2013 dataset while DP-LinkNet performed best on the DIBCO2017 dataset. The 2-StageGAN performed best on the DIBCO2018 dataset while SauvolaNet outperformed the others on the DIBCO2019 challenge.
Finally, we make the code, all models and evaluation publicly available\footnote{\url{https://github.com/RichSu95/Document_Binarization_Collection}} to ensure reproducibility and simplify future binarization evaluations. 
\end{abstract}
\section{Introduction}
Image binarization is a process that converts a color or grayscale image into an image whose pixels can have only two different values, usually black and white. In the domain of document image analysis, binarization typically consists in separating the text (foreground) from its support (background), \eg the paper.
While it became less popular for text recognition, it remains an important pre-processing step in many other tasks, such as writer identification~\cite{Christlein17PR,Christlein17ICDAR}, word spotting or optical character recognition (OCR)~\cite{sauvolanet}.
In traditional global binarization, the grayscale intensity frequency histogram of an image is analyzed and an appropriate threshold is set, \eg Otsu's thresholding~\cite{Otsu}. 
Alternatively, binarization is applied locally using statistics such as mean and standard deviation like the popular Sauvola method~\cite{SauvolaAlgo}.
However, these methods have problems with ink bleed-through artifacts and other %
artifacts such as stains, blurring, faint characters and noise~\cite{Mustafa}. An error that may be generated through incorrect binarization may propagate forward and lead to performance reduction in subsequent tasks. 
Document binarization also acts as a means to filter out these undesirable features. 
A thorough overview of binarization techniques, datasets, and metrics is given in a survey by Tensmeyer and Martinez~\cite{Tensmeyer}.

In recent years, rather than relying on traditional image binarization techniques, many studies have been conducted that employ deep learning models to binarize document images. The advent of deep learning has brought a multitude of changes to the domain of computer vision and image processing. Convolutional neural networks (CNNs) identify features automatically by learning from training data. The image features are discovered at multiple layers and are learned gradually from lower-levels to higher-levels. This multi-layered architecture performs a series of convolutions on the input image. A training process is implemented to adjust the parameters of the network to achieve the desired output.

In the past decade, there have been immense progress in the field of binarization of contemporary and historical documents using deep learning techniques. Although many approaches using deep learning for document binarization have been put forward, it is difficult to identify which among these models perform best when compared to one another.
The root cause of this problem is the fact that most of these models have never been trained and tested on a common dataset using the same evaluation protocol. 
This paper aims to resolve this disparity by training and testing some well-known binarization models~\cite{degan,sauvolanet,robin,deepotsu,twoStage,sae,dplink} on common datasets from the well-known Document Image Binarization Contests (DIBCO)~\cite{dibco2009,dibco2010,dibco2011,dibco2012,dibco2013,dibco2014,dibco2016,dibco2017,dibco2018,dibco2019}.
While we evaluated the results of the models using four metrics, we omitted investigations on the relationship between result quality and processing time as Lins \textit{et~al}.\ did~\cite{lins2021icdar}.
Our evaluations draw a very heterogeneous picture. All four evaluation datasets have a different winner. Overall, DE-GAN ranks best across the four chosen DIBCO test datasets while metric-wise, the 2-Stage GAN outperforms the other models. 

The following section~\ref{overview} of the paper provides a brief overview on the network architectures and methodologies used in the different binarization models that would be compared against one another. Section~\ref{material} gives a detailed description on the various datasets, validation metrics and on how all the models were trained. Section~\ref{results} shows the results of evaluating all models on the various test datasets and provides a brief discussion on the outcome of the experiments. 

\section{Overview of Evaluated Binarization Methods} \label{overview}
\subsection{Document Enhancement Generative Adversarial Network}
The work presented by Souibgui \textit{et al.}~\cite{degan} models the document binarization problem as an image-to-image translation task. The Document Enhancement Generative Adversarial Network (DE-GAN) model basically consists of a generator and a discriminator. The generator follows a U-Net architecture~\cite{unetpaper} and its objective is to generate a clean image given the original degraded image. The goal of the discriminator is then to determine if the image shown is a fake image generated by the generator or the original binarized ground truth. An adversarial loss function is employed for training the model~\cite{degan}: 

\begin{equation}
\begin{split}
    L_{GAN}(\phi_{G}, \phi_{D}) &= E_{I^{W}, I^{GT}} log[D_{\phi D}(I^{W},I^{GT})]\\
    &+ E_{I^{W}, I^{GT}} log[1 - D_{\phi D}(I^{W},G_{\phi G}(I^{W}))],
\end{split}
\label{eq:deganloss1}
\end{equation}

where $G_{\phi G}$ and $D_{\phi D}$ are the generator and discriminator functions respectively, $I^{W}$ is the degraded image and $I^{GT}$ is the ground truth. 
After a few epochs, the network is able to generate images similar to the ground truth. 
To maintain a good text quality and to improve training speed an additional log loss function is added.
The objective is that the text output from the generator is identical to the ground truth text~\cite{degan}: 

\begin{equation}
\begin{aligned}
   L_\text{log}(\phi G) = E_{I^{GT}, I^{W}}[-(I^{GT}\log(G_{\phi G}(I^W)) + ((1-I^{GT})\log(1-G_{\phi G}(I^W)))].
\end{aligned}
\label{eq:deganloss2}
\end{equation}
The overall loss of the network is denoted as~\cite{degan}:

\begin{equation}
\begin{aligned}
   L_\text{net}(\phi G, \phi D) = \min_{\phi G}\max_{\phi D}L_\text{GAN}(\phi G, \phi D) + \lambda L_\text{log}(\phi G),
\end{aligned}
\label{eq:deganloss3}
\end{equation}
where $L_\text{GAN}$ is the adversarial loss function used to train the cGAN and $\lambda$ is a hyper-parameter that is set to 500 for document binarization. The generator follows an encoder-decoder structure. The encoder performs down-sampling of the given input up to a certain layer and the decoder then up-samples the encoder output. The discriminator used is a simple Fully-Connected Network (FCN) with 6 convolutional layers. To train the DE-GAN model, overlapped patches of size \numproduct{256 x 256} pixel are obtained from the degraded images and fed as input to the generator.

\subsection{SauvolaNet}
Inspired by the traditional Sauvola thresholding algorithm~\cite{SauvolaAlgo}, the work by Li \textit{et~al.}~\cite{sauvolanet} presents a deep learning approach to learn the Sauvola parameters, called the ``SauvolaNet''. The network aims to making the model computationally efficient. %
The model also comprises of an attention mechanism that aims to estimate the required Sauvola window sizes for each pixel location. One main drawback of the traditional Sauvola thresholding approach is that the algorithm achieves its highest performance only when the right hyperparameters are manually tuned for each input image (window size, estimated level of document degradation and dynamic range of input image intensity). SauvolaNet uses three modules, the Multi-Window Sauvola (MWS), Pixelwise Window Attention (PWA), and Adaptive Sauvola Threshold (AST) to learn an auxiliary threshold estimation function.

The MWS module takes an image as input and uses the Sauvola algorithm to estimate the local thresholds for different window sizes. The PSA module also takes the same image as input to estimate the window sizes for each pixel location. The AST module then predicts the final threshold for each pixel location by fusing the thresholds of different windows from the MWS and weights from the PWA modules. The SauvolaNet function is modelled as~\cite{sauvolanet}:

\begin{equation}
\begin{aligned}
    T = g_{SauvolaNet}(D),
\end{aligned}
\label{eq:sauvola1}
\end{equation}

where, $T$ is the output, $g_{SauvolaNet}$ is the auxiliary threshold estimation function and $D$ is the input image. The PWA uses instance normalization instead of batch normalization in order to avoid overfitting when training with a small dataset. When training the SauvolaNet, the input image $D$ is normalized to values in the range (0,1) and a modified hinge loss was developed~\cite{sauvolanet}:

\begin{equation}
\begin{aligned}
    \text{loss}[i,j] = \max (1-\alpha\cdot (D[i,j] - T[i,j]) \cdot B[i,j],0),
\end{aligned}
\label{eq:sauvola2}
\end{equation}
where $B$ is the binarization ground truth with values -1 for foreground and +1 for the background. $i$ and $j$ are indices that specify the location of a pixel. $\alpha$ is a parameter to control the margin of the decision boundary and only the pixels close to the decision boundary are used in gradient-backpropagation.

\subsection{Two-Stage GAN}
The work presented by Suh \textit{et al.} proposes a two-stage color document image binarization deep learning architecture using generative adversarial neural networks (GANs)~\cite{twoStage}. The GAN architecture generally consists of two networks, i.e., the generator and the discriminator. For this model, the EfficientNet~\cite{efficientnet} was used as the generator on account of its efficiency in the domain of image classification. In the case of the discriminator, the discriminator network from the PatchGAN~\cite{patchgan} was implemented.

The first part of the network consists of four color independent generators that are trained with the red, green, blue, and gray channels in order to generate an enhanced image by removing background color information. The resulting channel images and corresponding ground truths first concatenated and then fed to the discriminator network. The binarization in the first stage is performed using local predictions in small patches. In order to cater to regions with larger backgrounds, the second stage of the network performs global binarization with the resized original input image and local binarization using the first stage output. Except for the input image channels, the structure for the generators in the second stage is identical to that of the first stage. During training, the images are divided into patches of 256 $\times$ 256 pixels resolution without scaling. When training GANs in general, it is common to observe an instability in loss function convergence~\cite{twoStage}. To solve this issue, the Wasserstein GAN with penalty was used which implements the Wasserstein K-distance as the loss function. Further, instead of the typically used L1 loss, pixel-wise binary cross-entropy is defined as the additional loss term for the generator update.

\subsection{Robin U-Net model}
The implementation by Mikhail Masyagin~\cite{robin} presents the Robust Documentation Binarization (ROBIN) tool. ROBIN makes use of a simple U-Net model~\cite{unetpaper} to perform document binarization. The U-Net model was originally developed for the purpose of semantic segmentation of medical images. The U-Net architecture can be described as an encoder-decoder network. The input image is first fed into the encoder network, where multiple convolution blocks are applied followed by a maxpool downsampling layer. The idea here is to encode the input image into feature representations at multiple levels. The output from the encoder is then sent to the decoder where the activation map undergoes upsampling or deconvolution. Skip connections are also introduced between the encoder and decoder structure such that the deep and shallow features can be combined. 

When training the model, the input images are split into patches of $128 \times 128$\,px resolution. The learning rate was set to 0.0001 with the Adam optimizer. 
The training is trained using the dice coefficient loss and run for 250 epochs with an early stopping criteria.

\subsection{DP-LinkNet}
The DP-LinkNet is a segmentation model introduced by Xiong \textit{et al.}. It makes use of the D-LinkNet~\cite{dlinknet} and LinkNet~\cite{linknet} models with a pre-trained encoder as the backbone. 

The model consists of: 1) an encoder, 2) a hybrid dilated convolution module, 3) a spatial pyramid pooling (SPP) module, and 4) a decoder~\cite{dplink}. Firstly, the input image is fed to the encoder where the text stroke features are extracted. The series of convolutions and down-sampling occurring at the encoder causes a reduction in the resolution of the obtained feature map. To counter this effect, dilated convolutions are introduced into the model. Dilated convolutions help in exponentially increasing the size of the receptive field without affecting the spatial resolution. An issue that still persists here is the fact that the dilated convolution module may still find it difficult to identify objects of different sizes with a fixed-sized field-of-view. To counter this effect, the spatial pyramid pooling is employed. This helps to present the input feature maps at different scales. Lastly, the decoder performs transposed convolution. %
Skip connections between the decoder and encoder structure are present to combine the shallow-level and high-level features, helping to compensate any loss encountered by convolution and pooling operations. When training the model, the binary cross entropy and dice coefficient losses are used. The input images were split into patches of size $128 \times 128$ px. The adam optimizer was set with an initial learning rate of $2 \times 10^{-4}$. The model was trained for 500 epochs with an early stopping criteria to avoid overfitting.

\subsection{Selectional Auto-Encoder}

The work presented by Calvo-Zaragoza \textit{et al.}~\cite{sae} uses an auto-encoder network topology to perform an image-to-image processing task. Such a task results in higher computational efficiency since all pixels in the input image are processed at the same time. Generally, an auto-encoder network is trained to learn the identity function. However, in the selectional auto-encoder (SAE), the network is trained to learn a selectional map over a $w \times h$ image, preserving the input shape. The activation of each pixel depends on whether the pixel belongs to the foreground or the background. When training the SAE, the images along with their corresponding ground-truth (binarized image) are fed as input to the network. Auto-encoders are feed-forward networks and generally consist of two sections, i.e., the encoder and decoder. The encoder learns to extract the latent representation given an input image, downsampling the image until an intermediate representation is achieved. The output from the encoder is then upsampled and reconstructed to the original input image dimensions by the hidden layers of the decoder. The last layer consists of a set of neurons and a sigmoid activation layer which then gives an output prediction between the range of 0 and 1. 

Since the binarized output image should consist of pixel values being 0 or 1 and not in between, a thresholding process is implemented to decide whether the certain pixel belongs to the background or foreground. The encoder and decoder both consisted of 5 layers each and the sampling operators were fixed at $2 \times 2$. Network weights were initialized using Xavier initialization~\cite{Xavier}. Optimization is handled with stochastic gradient descent and a mini-batch size of 10. The initial learning rate is set to 0.001 and the network is trained for 200 epochs with an early stopping criteria kept in place.

\subsection{DeepOtsu}
The work presented by He \textit{et al.}~\cite{deepotsu} proposes an iterative deep learning approach to obtain binarized images called the DeepOtsu model. However, unlike the aforementioned methods in this section, the deep learning network in this case aims to remove artifacts and generate a non-degraded version of the input image. The degraded input image $\bf{x}$ is modeled as: 
\begin{equation}
\begin{aligned}
    \bf{x} = \bf{x}_\text{u} + e,
\end{aligned}
\label{eq:deepotsu}
\end{equation}

where $\bf{x}_\text{u}$ is the latent uniform image and $e$ is the degradation. The aim of the deep learning network is to ultimately obtain $\bf{x}_\text{u}$. 

The network was trained with images split into patches of size $256 \times 256$. The patches are first fed to the CNN model and the obtained output is then compared to the ground truth, which in this case should be representative of the uniform, clean version of the input image. To obtain this ground truth, the degraded input image is compared to the already available binarized images from the dataset. Then, the ground truth image is computed as the average pixel value with the same label within the image patch. Once the non-degraded, uniform version of the input image is obtained, the binarized version of the image can be easily obtained using Otsu thresholding~\cite{Otsu}. The basic U-Net model~\cite{unetpaper} is used for learning the degradation. The down-sampling path of the network consisted of 5 convolutional layers with a $3 \times 3$ kernel size, followed by a leaky-ReLU activation~\cite{LeakyRelu} and $2 \times 2$ max pooling. The batch size was set to 8 and the learning rate set to $10^{-4}$.

\section{Materials and Methods} \label{material}
\subsection{Datasets}
\begin{figure}[t]
    \centering
    \begin{subfigure}[b]{0.45\textwidth}
        \centering
        \includegraphics[width=0.9\textwidth]{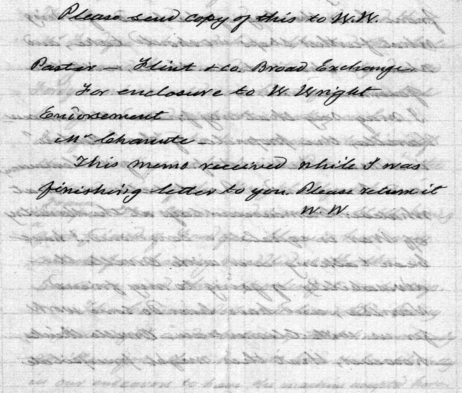}
        \caption{DIBCO2013}
        \label{fig:DIBCO2013}
    \end{subfigure}
    \qquad
    \begin{subfigure}[b]{0.45\textwidth}  
        \centering 
        \includegraphics[width=0.9\textwidth]{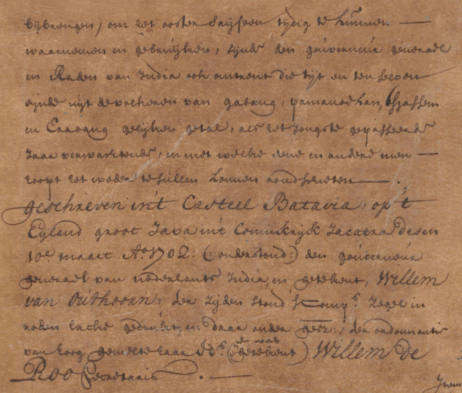}
        \caption{\small DIBCO2017}
        \label{fig:DIBCO2017}
    \end{subfigure}
    \vskip\baselineskip
    \begin{subfigure}[b]{0.45\textwidth}   
        \centering 
        \includegraphics[width=0.9\textwidth]{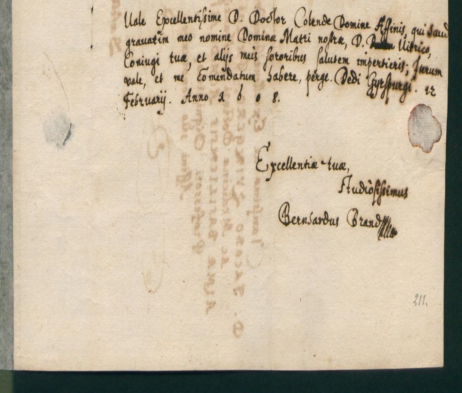}
        \caption{DIBCO2018}
        \label{fig:DIBCO2018}
    \end{subfigure}
    \qquad
    \begin{subfigure}[b]{0.45\textwidth}   
        \centering 
        \includegraphics[width=0.9\textwidth]{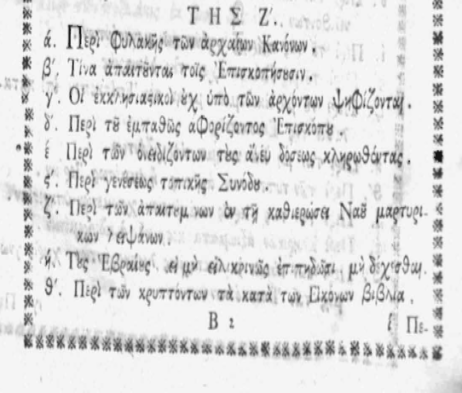}
        \caption{DIBCO2019}
        \label{fig:DIBCO2019}
    \end{subfigure}
    \caption{Image examples from the different DIBCO datasets used for testing the models.} 
    \label{fig:DIBCO examples}
\end{figure}

All models mentioned in the previous section are trained and tested on document images from the DIBCO dataset. To keep the comparison between the models fair and precise, the training set and validation set remain the same for all models. The training set consists of the DIBCO2009, DIBCO2010, DIBCO2011, DIBCO2012, DIBCO2014, and DIBCO2016 datasets. The models are evaluated on DIBCO2013, DIBCO2017, DIBCO2018, and DIBCO2019 datasets. The four test sets were chosen based on the unique properties present in the three sets. DIBCO2013 consists of both handwritten and printed documents. The images from DIBCO2017 had more textual content in them. The DIBCO2018 dataset consisted of images of textual content present towards the borders or corners of the papers and higher intensity of bleed-through artifacts. 
The DIBCO2019 dataset had large variations in the types of images. Note that we used only track A since track B, containing text content on papyri, are not present in any training data which lead to rather poor learning-based results. 
Evaluations based on these four datasets give an idea of how well the models are able to generalize on different types of unseen images. Figure~\ref{fig:DIBCO examples} shows some samples of images that belong to the DIBCO datasets used for validating the models.

\subsection{Metrics}
Our evaluation of the various models is based on the standard evaluation metrics used in the DIBCO challenges: (1)~F-measure (FM), (2)~pseudo F-measure (pFM), (3)~peak signal to noise ratio (PSNR), and (4)~distance reciprocal distortion (DRD).
The FM and pFM reach their best value at 1 and worst at 0 (\cref{eq:fm,eq:pfm}).
PSNR describes how close the binarized and ground truth images are (\cref{eq:psnr}). The higher the PSNR, the better is the binarized result. The DRD is based on the reciprocal of distance, matching well to subjective evaluation by human visual perception (\cref{eq:drd}).

\begin{equation}
\begin{aligned}
    \text{FM} =  \frac{2 \times \text{Recall} \times \text{Precision}}{\text{Recall} + \text{Precision}},
\end{aligned}
\label{eq:fm}
\end{equation}

where, $\text{Recall} = \frac{TP}{TP + FN}$ and $\text{Precision} = \frac{TP}{TP + FP}$. TP, FP and FN denote true positive, false positive and false negative values respectively.

\begin{equation}
\begin{aligned}
    \text{pFM} =  \frac{2 \times \text{pRecall} \times \text{pPrecision}}{\text{pRecall} + \text{pPrecision}},
\end{aligned}
\label{eq:pfm}
\end{equation}

where, pRecall and pPrecision, respectively the pseudo-recall and the pseudo-precision, are metrics weighted based on the distance to the contours of the foreground in the ground truth.
For the pseudo-recall, pixels around strokes have weights starting from 1, and reaching 0 at a distance corresponding to the stroke's width, and pixels inside of the strokes have a weight of 1.
For the pseudo-precision, pixels outside strokes but not further than the stroke's thickness have a weight of 1, and inside the stroke their weight increase toward the center, where they reach a value of 2.

\begin{equation}
\begin{aligned}
    \text{PSNR} = \log_{10} \left(\frac{C^2}{\text{MSE}}\right),
\end{aligned}
\label{eq:psnr}
\end{equation}

where, $\text{MSE} = \frac{ \sum\nolimits_{x=1}^{m}  \sum\nolimits_{y=1}^{n} (L(x,y) - L'(x,y))^2}{mn}$. The terms $m$ and $n$ denote the dimensions of the image. $C$ denotes the difference present between the text and background.

\begin{equation}
\begin{aligned}
    \text{DRD} = \frac{ \sum_{k} \text{DRD}_k }{\text{NUBN}},
\end{aligned}
\label{eq:drd}
\end{equation}

where $\text{DRD}_k$ is the distortion of the \textit{k}th flipped pixel and $\text{NUBN}$ is the number of non-uniform $8 \times 8$ blocks in the ground truth image. 

\begin{figure}[t!]
    \centering
    \begin{subfigure}[b]{0.32\textwidth}
        \centering
        \includegraphics[width=0.9\textwidth]{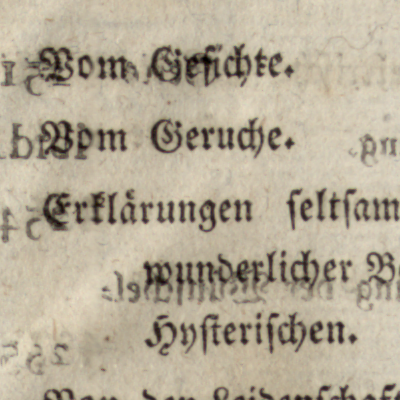}
        \caption{Input image}
        \label{fig:X1}
    \end{subfigure}
    \hfill
    \begin{subfigure}[b]{0.32\textwidth}  
        \centering 
        \includegraphics[width=0.9\textwidth]{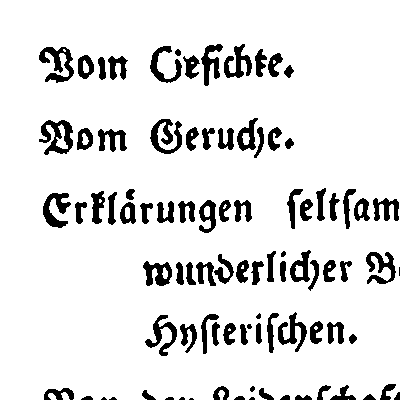}
        \caption{Ground Truth}
        \label{fig:X2}
    \end{subfigure}
    \hfill
    \begin{subfigure}[b]{0.32\textwidth}  
        \centering 
        \includegraphics[width=0.9\textwidth]{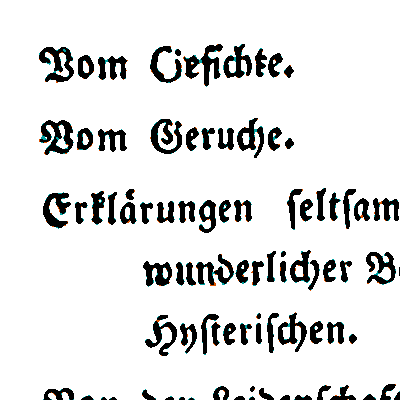}
        \caption{DE-GAN}
        \label{fig:X3}
    \end{subfigure}
    \vskip\baselineskip
    \begin{subfigure}[b]{0.32\textwidth}
        \centering
        \includegraphics[width=0.9\textwidth]{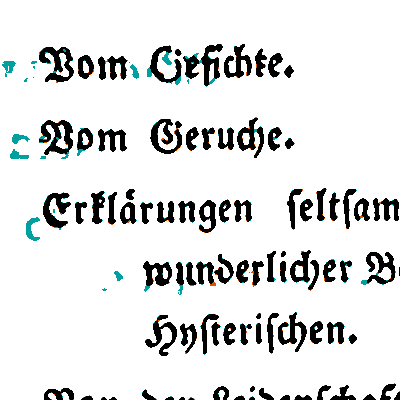}
        \caption{Robin (U-Net)}
        \label{fig:X4}
    \end{subfigure}
    \hfill
    \begin{subfigure}[b]{0.32\textwidth}  
        \centering 
        \includegraphics[width=0.9\textwidth]{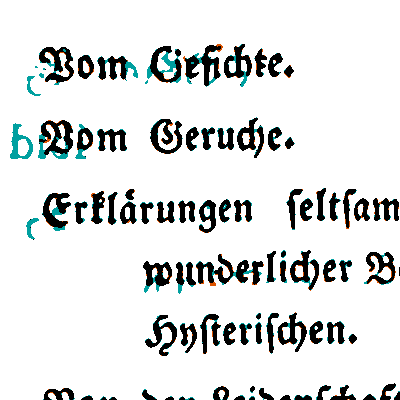}
        \caption{DeepOtsu}
        \label{fig:X5}
    \end{subfigure}
    \hfill
    \begin{subfigure}[b]{0.32\textwidth}  
        \centering 
        \includegraphics[width=0.9\textwidth]{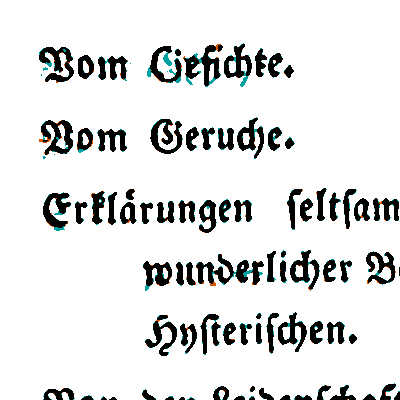}
        \caption{2-Stage GAN}
        \label{fig:X6}
    \end{subfigure}
    \vskip\baselineskip
    \begin{subfigure}[b]{0.32\textwidth}
        \centering
        \includegraphics[width=0.9\textwidth]{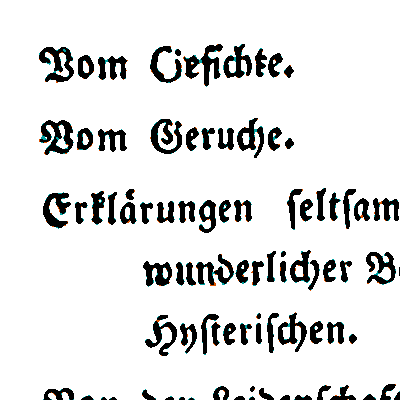}
        \caption{DP-LinkNet}
        \label{fig:X7}
    \end{subfigure}
    \hfill
    \begin{subfigure}[b]{0.32\textwidth}  
        \centering 
        \includegraphics[width=0.9\textwidth]{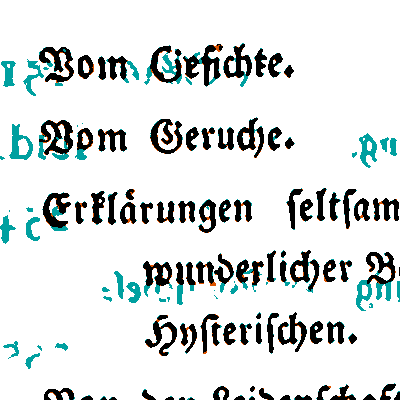}
        \caption{SAE}
        \label{fig:X8}
    \end{subfigure}
    \hfill
    \begin{subfigure}[b]{0.32\textwidth}  
        \centering 
        \includegraphics[width=0.9\textwidth]{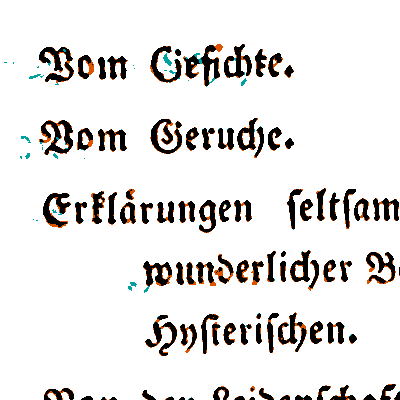}
        \caption{SauvolaNet}
        \label{fig:X9}
    \end{subfigure}
    \caption{Illustration of some results for an image from DIBCO-2017. Pixels in cyan are false positives. The few pixels in orange are false negatives. Pixels in white or black match the ground truth.} 
    \label{fig:result-examples}
\end{figure}

\subsection{Training}
All models are trained on the DIBCO datasets as mentioned in the previous sections. Based on the configuration of the models, the degraded images along with the accompanying ground truths are first split into patches of size \numproduct{256 x 256} pixel or \numproduct{128 x 128} pixel resolutions. The patches are further augmented by random horizontal flipping, vertical flipping and rotations. %
The number of epochs for training each model is set based on the recommendation of the authors for each model, along with an early stopping criteria to monitor any possibility of overfitting the models. 
If the validation loss of the model does not show significant changes for 15 consecutive epochs, the training would stop and the model would be saved. Certain pre-processing and post-processing operations on the images exclusive to specific models have also been implemented. Such an example is the application of Otsu's thresholding on the output of the DeepOtsu method. The hyper-parameters for the models are optimized using the python library ``optuna''~\cite{optuna}.

\begin{table}[t]
\centering
\caption{Results of different image binarization methods on the \subref{table:results2013} DIBCO2013, \subref{table:results2017} DIBCO2017, \subref{table:results2018} DIBCO2018, and \subref{table:results2019} DIBCO2019 datasets.
Note that the winners of the respective DIBCO2017, DIBCO2018 and DIBCO2019 challenge had more data available.
}
\begin{subtable}{0.5\textwidth}
\centering
\begin{tabular}{l@{\hspace{1.5em}}cccc} 
\toprule
Model         & PSNR\textuparrow           & FM\textuparrow            & pFM\textuparrow           & DRD\textdownarrow            \\ 
\midrule
DE-GAN        & \textbf{24.08} & \textbf{97.68} & \textbf{98.09} & \textbf{1.11}  \\
Robin (U-Net) & 22.81          & 95.07         & 95.82         & 1.99           \\
DeepOtsu      & 21.19          & 93.46         & 95.99         & 2.25           \\
2-Stage GAN   & 22.60          & 95.75         & 96.40         & 1.46           \\
DP-LinkNet    & 23.63          & 96.49         & 97.24         & 1.10           \\
SAE           & 20.88          & 93.35         & 94.44         & 3.17           \\
SauvolaNet    & 23.41          & 96.31         & 97.53         & 1.28           \\
\midrule
Winner~\cite{dibco2013,dibco2017} & 20.68 & 92.12 & 94.19 & 3.10\\
\bottomrule
\end{tabular}
\caption{DIBCO2013}
\label{table:results2013}
\end{subtable}
\qquad
\begin{subtable}{0.37\textwidth}
\centering
\begin{tabular}{cccc}
\toprule
PSNR\textuparrow & FM\textuparrow & pFM\textuparrow & DRD\textdownarrow\\
\midrule
18.31          & 96.23         & \textbf{98.10}& 3.22          \\
19.99          & 92.05         & 94.06         & 2.23          \\
18.02          & 89.01         & 91.84         & 3.50          \\
20.89          & 95.56         & 96.54         & 1.33          \\
\textbf{22.84} & \textbf{97.92}& 97.94         & \textbf{0.77} \\
16.73          & 87.59         & 90.41         & 5.60          \\
19.40          & 93.33         & 96.26         & 2.20          \\
\midrule
18.28 & 91.04 & 92.86 & 3.40\\
\bottomrule
\end{tabular}
\caption{DIBCO2017}
\label{table:results2017}
\end{subtable}

\begin{subtable}{0.5\textwidth}
\centering
\begin{tabular}{l@{\hspace{1.5em}}cccc}
\toprule
Model & PSNR\textuparrow & FM\textuparrow & pFM\textuparrow & DRD\textdownarrow \\
\midrule
DE-GAN         & 15.98          & 76.21         & 83.29         & \phantom{0}8.01          \\
Robin (U-Net)  & 15.78          & 78.80         & 81.11         & 12.20         \\
DeepOtsu       & 12.72          & 66.60         & 68.83         & 42.52         \\
2-Stage GAN    & \textbf{19.93} & \textbf{92.40}& \textbf{94.90}& \textbf{\phantom{0}2.67} \\
DP-LinkNet     & 15.73          & 78.56         & 80.70         & 13.72         \\
SAE            & 14.48          & 73.45         & 76.33         & 15.45         \\
SauvolaNet     & 16.03          & 77.94         & 81.92         & 10.41         \\ 
\midrule
Winner~\cite{dibco2018,dibco2019} & 19.11 & 88.34 & 90.24 & 4.92\\
\bottomrule
\end{tabular}
\caption{DIBCO2018}
\label{table:results2018}
\end{subtable}
\qquad
\begin{subtable}{0.37\textwidth}
\centering
\begin{tabular}{cccc} 
\toprule
PSNR\textuparrow           & FM\textuparrow             & pFM\textuparrow           & DRD\textdownarrow            \\ 
\midrule
15.12           & 70.86          & 70.69        & \phantom{0}6.23            \\
14.39          & 65.55          & 65.34          & \phantom{0}7.36           \\
14.82          & 70.81          & 70.91          & \phantom{0}7.59           \\
12.87          & 65.09          & 65.72          & 12.71          \\
14.20          & 61.84          & 61.55          & \phantom{0}7.58           \\
12.50          & 62.17          & 61.90           & 13.43          \\
\textbf{15.83} & \textbf{72.04} & \textbf{71.59} & \textbf{\phantom{0}5.55}  \\
\midrule
14.48 & 72.88 & 72.15 & 16.24\\
\bottomrule
\end{tabular}
\caption{DIBCO2019}
\label{table:results2019}
\end{subtable}
\end{table}

\section{Evaluation} \label{results}
The results of testing each model on the different test DIBCO datasets are as shown in the following tables. \Cref{table:results2013} shows the results of validating the models on the DIBCO2013 dataset. The DIBCO2013 dataset contains images that have a good representation of the training data, without any major artifacts or degradation present. All methods display comparable performance with the DE-GAN performing best.
For reference, we also show the DIBCO winners of the respective challenge. Note that the participants of 2017 and later potentially used more data for training. 

\Cref{table:results2017} shows the results of validating the models on the DIBCO2017 dataset. Here, the performance of the models start to fluctuate more when compared to \cref{table:results2013}. This might be due to the fact that the DIBCO2017 dataset contains more images that have more densely packed textual content. The DP-LinkNet model outperforms the other models in terms of PSNR, FM and DRD whereas the DE-GAN model has a higher performance in terms of pFM. However, it can be observed that the DRD value for DE-GAN is quite high, indicating that the resulting binarized images have higher rate of distortions. This may be attributed to the training process of the DE-GAN model, which may have introduced distortions to the generated images.
Qualitative results for a randomly chosen sample from DIBCO2017 can be seen in \cref{fig:result-examples}.

\begin{figure}[t!]
	\centering
	\subcaptionbox{\label{fig:orig_border}}{
		\includegraphics[width=0.4\textwidth]{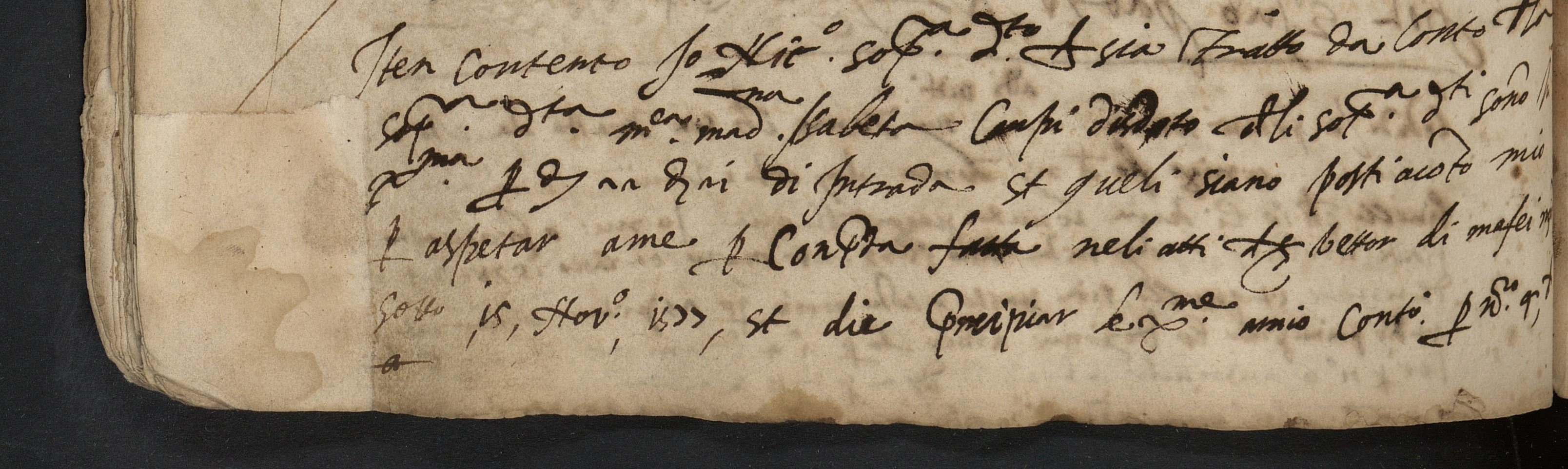}
	}
	\quad
	\subcaptionbox{\label{fig:fail_dplinknet}}{
		\includegraphics[width=0.4\textwidth]{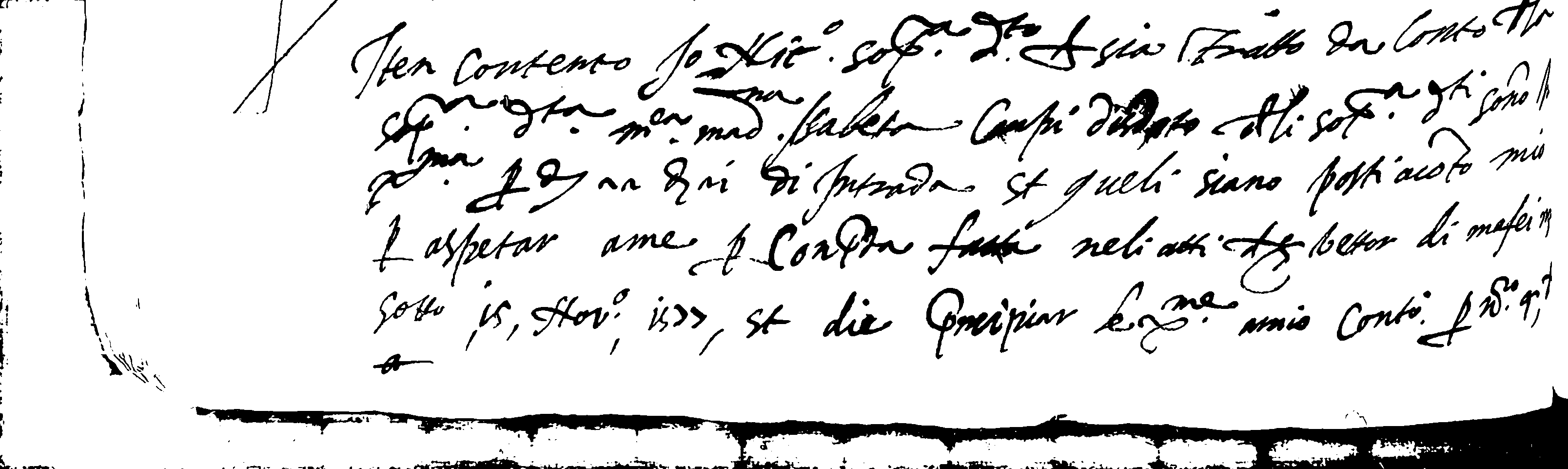}
	}

	\subcaptionbox{\label{fig:orig_squared_paper}}{
		\includegraphics[width=0.4\textwidth]{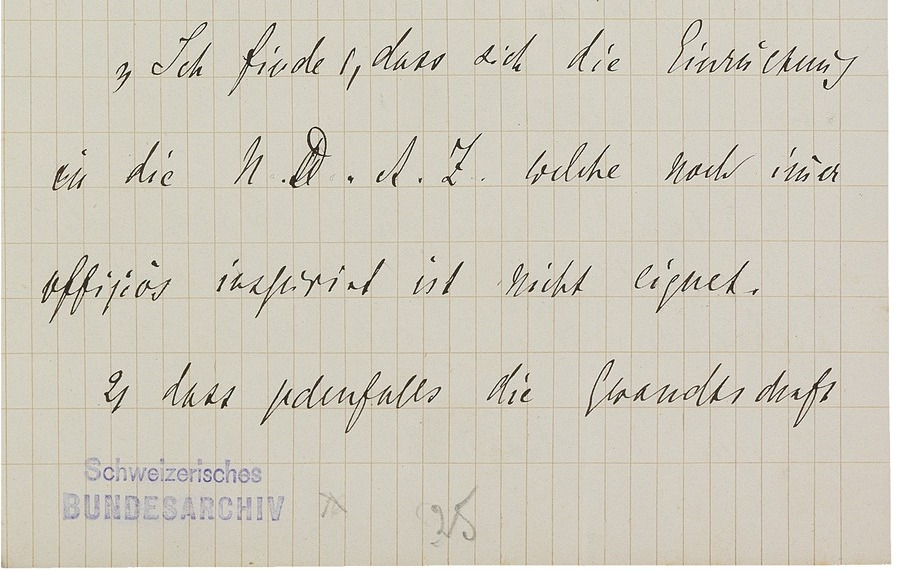}
	}
	\quad
	\subcaptionbox{\label{fig:fail_2-stage_gan}}{
		\includegraphics[width=0.4\textwidth]{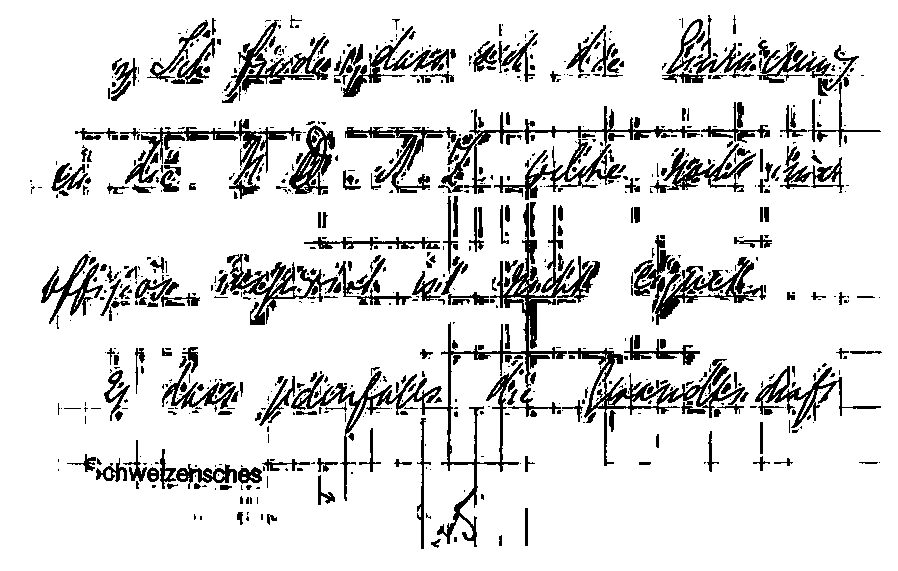}
	}

	\caption{Qualitative examples of failure modes: 
		\subref{fig:fail_dplinknet} shows that DP-LinkNet binarizes the large black
		borders present in images of DIBCO2018
		to white;
		\subref{fig:fail_2-stage_gan} shows that the 2-Stage GAN struggles with the
		squared paper given in images of DIBCO2019, and additionally produces halo-artifacts. 
	}
\end{figure}

The results for the DIBCO2018 dataset is shown in \cref{table:results2018}. The winner is clearly the 2-Stage GAN approach, outperforming all other methods in each metric. 
For the pFM and the DRD metrics, the DE-GAN ranks second. 
Interestingly, the DP-LinkNet struggles with black page borders, see \cref{fig:fail_dplinknet}. 
While it wins for the 2017 dataset that does not have borders, it performed poorly on images that have borders that are present in the DIBCO2018 dataset, \cf \cref{fig:DIBCO2018}.

While SauvolaNet ranks behind these two methods in the DIBCO2018 challenge, it outperforms both methods on the DIBCO2019 dataset, see \cref{table:results2019}. 
The 2-Stage GAN, which performs very well for the 2013 to 2018 datasets had some difficulties to deal with the squared paper (check paper, quadrille paper) of the 2019 dataset, which can be observed in \cref{fig:fail_2-stage_gan}.
\begin{table}[t!]
\centering
\caption{Average over \subref{tab:avg_metrics} all metrics and \subref{tab:avg_ranks} all ranks. Runtimes evaluated using an NVIDIA RTX 2060 GPU (12 GB RAM). Note that DeepOtsu and 2-Stage GAN were limited by the available memory.}
\label{tab:avg}
\begin{subtable}{0.7\textwidth}
    \centering
\begin{tabular}{l@{\hspace{1.5em}}ccccc}
\toprule
Model & PSNR\textuparrow & FM\textuparrow & pFM\textuparrow & DRD\textdownarrow & img/sec\textuparrow\\
\midrule
DE-GAN          &18.37  &85.25  &87.54  &\phantom{0}4.64 & 0.67\\
Robin (U-Net)   &18.24  &82.87  &84.08  &\phantom{0}5.95 & 1.99\\
DeepOtsu        &16.69  &79.97  &81.89  & 13.96 & 0.01\\
2-Stage GAN     &19.07  &\textbf{87.20}  &\textbf{88.39}  &\textbf{\phantom{0}4.54} & 0.01\\
DP-LinkNet      &\textbf{19.10}  &83.70  &84.36  &\phantom{0}5.79 & 0.49\\
SAE             &16.15  &79.14  &80.77  &\phantom{0}9.41 & 0.68\\
SauvolaNet      &18.67  &84.91  &86.83  &\phantom{0}4.86 & 0.37\\
\bottomrule
\end{tabular}
    \caption{Average metrics}
    \label{tab:avg_metrics}
    \end{subtable}
    \qquad
    \begin{subtable}{0.2\textwidth}
    \centering
    \begin{tabular}{c}
    \toprule
    Avg.\ rank\textdownarrow\\
        \midrule
\textbf{2.44}\\
4.19\\
5.50\\
3.25\\
3.38\\
6.63\\
2.63\\
         \bottomrule
    \end{tabular}
    \caption{Average ranks}
    \label{tab:avg_ranks}
\end{subtable}
\end{table}
When we average all metrics for all different evaluated datasets, see \cref{tab:avg_metrics}, the 2-Stage GAN seems to be on average the most suitable binarization method appearing to be consistent in terms of performance. 
Interestingly, computing the average rank over all metrics, \ie the average over all 16 ranks for each method, it falls behind DE-GAN and SauvolaNet, \cf \cref{tab:avg_ranks}. 

We also evaluated the runtime, reported as throughput, \ie images per second in the last column of \cref{tab:avg_metrics}. The best throughput has the Robin binarization method. Note, however that we evaluated the methods on a small-sized GPU (NVIDIA RTX 2060) with 12\,GB GPU-RAM. Unfortunately, this affected the throughput of DeepOtsu and 2-Stage GAN because multiple images of the DIBCO 2013 dataset contain very large images, \eg image sizes of $4161 \times 1049$.

\section{Conclusion}
In this paper, we thoroughly evaluated seven deep learning-based methods in a fair evaluation where we fixed the data and augmentation used. 
We evaluated the methods using all ten available DIBCO datasets. 
Therefore, we used six datasets for training and the remaining four datasets for testing. 
Our evaluations show that the results are very diverse on the four different tested datasets and no clear winner could be established. 
Overall, the DE-GAN approach achieved the best rank averaged over all four different datasets followed by SauvolaNet. 
When we compare the metrics individually, then the 2-Stage GAN approach performed best followed by the DE-GAN. 
In the very different DIBCO2019 dataset, however, the SauvolaNet outperformed these methods. 

For future work, we would like to evaluate the methods also with a different protocol.
In particular, we would like to simulate the DIBCO scenario of each year's challenge to be comparable with the single DIBCO papers, \ie training with the datasets 2015--2016, then evaluating with 2017, adding 2017 to the training set, re-train and evaluate on 2018, and so on.
The use of additional augmentation techniques as well as additional training datasets is also worth investigating and might have huge impact on the overall performance of the binarization methods. 
Furthermore, pixel-based evaluation is not optimal~\cite{Tensmeyer}.
While the pFM metric incorporates the distance to the script contour, it might be worth investigating indirect measures, such as OCR/HTR accuracy or purely skeleton-based metrics~\cite{Silva20}.
From a practical point of view, the inference time is also worth investigating. This has been mainly studied in the competitions on time-quality document image binarization

\bibliographystyle{splncs04}
\bibliography{doc_lit}
\end{sloppypar}
\end{document}